\documentclass{article} 
\usepackage{nips15submit_e,times}
\usepackage{hyperref}
\usepackage{url}
\usepackage{amsmath}
\usepackage{graphicx}
\usepackage{amssymb}
\usepackage{algorithm}
\usepackage{enumerate}

\title{Learning Gaussian Graphical Models Using Discriminated Hub Graphical Lasso}

\author{
Zhen Li, Jingtian Bai, Weilian Zhou \\
Department of Statistics\\
NC State University\\
Raleigh, NC 27695 \\
\texttt{zli34@ncsu.edu, jbai4@ncsu.edu, wzhou11@ncsu.edu} \\
}

\nipsfinalcopy 

\begin{document}

\maketitle

\begin{abstract}
We develop a new method called Discriminated Hub Graphical Lasso (DHGL) based on Hub Graphical Lasso (HGL) by providing prior information of hubs. We apply this new method in two situations: with known hubs and without known hubs. Then we compare DHGL with HGL using several measures of performance. When some hubs are known, we can always estimate the precision matrix better via DHGL than HGL. When no hubs are known, we use Graphical Lasso (GL) to provide information of hubs and find that the performance of DHGL will always be better than HGL if correct prior information is given and will seldom degenerate when the prior information is wrong.
\end{abstract}

\section{Introduction}

Graphical models have been used widely in various systems. A graph consists of nodes, representing variables, and edges between nodes, representing the conditional dependency between two variables \cite{Drton}. In order to get a sparse and interpretable estimate of the graph, many authors have considered the optimization problem in the form $\text{min}_{\mathcal{\mathbf{\Theta}} \in S}\{\ell(\mathbf{X},\mathbf{\Theta})+\lambda\left\|\mathbf{\Theta}-\text{diag}(\mathbf{\Theta})\right\|_1\}$,  where $\mathbf{X}$ is an $n \times p$ data matrix, $\mathbf{\Theta}$ is a $p\times p$ symmetric matrix which contains the variables of interest, and $\ell(\mathbf{X},\mathbf{\Theta})$ is a loss function \cite{Tan}. In many cases, we assume that $\ell(\mathbf{X},\mathbf{\Theta})$ is convex in $\bf \Theta$. In reality, there exists a specific kind of nodes called hubs \cite{Tan}, which are connected to a large number of other nodes. Tan et al. (2014) propose a method called Hub Graphical Lasso (HGL) to estimate the graph with hubs \cite{Tan}. They decompose $\bf \Theta$ as $\mathbf{Z}+\mathbf{V}+\mathbf{V}^T$, where $\mathbf{Z}$ represents conditional dependency between non-hub nodes and $\mathbf{V}$ represents conditional dependency between hub nodes and other nodes. Then they use Hub Penalty Function which is the minimal of the term below with respect to $\mathbf{V},\mathbf{Z}$ such that $\mathbf{\Theta} = \mathbf{Z} + \mathbf{V}+\mathbf{V}^T$ instead of the $l_1$ penalty in the above optimization problem
$$\lambda_1\left\|\mathbf{Z}-\text{diag}(\mathbf{Z})\right\|_1+\lambda_2\left\|\mathbf{V}-\text{diag}(\mathbf{V})\right\|_1+\lambda_3\left\|(\mathbf{V}-\text{diag}(\mathbf{V}))_j\right\|_q.$$

Since HGL does not require prior information of hubs and the number of hubs found by HGL is always not more than that of true hubs, we hope to advance HGL by providing prior information of hubs if we have some beforehand. We introduce Discriminated Hub Penalty Function with the form $\text{P}(\mathbf{\Theta})$, which is the minimal of the term below with respect to $\mathbf{V},\mathbf{Z}$ such that $\mathbf{\Theta} = \mathbf{Z}+\mathbf{V}+\mathbf{V}^T$
\begin{eqnarray}
\label{Eq:hubpenalty}
 \Bigg\{ \lambda_1 \| \mathbf{Z} - \text{diag}(\mathbf{Z})\|_1 +\lambda_2 \sum_{j\notin\mathcal{D}} \| (\mathbf{V}  - \text{diag}(\mathbf{V}))_j \|_1 + \lambda_3 \sum_{j\notin\mathcal{D}} \| (\mathbf{V}  - \text{diag}(\mathbf{V}))_j \|_q \nonumber \\
+ \lambda_4 \sum_{j\in\mathcal{D}} \| (\mathbf{V}  - \text{diag}(\mathbf{V}))_j \|_1 + \lambda_5 \sum_{j\in\mathcal{D}} \| (\mathbf{V}  - \text{diag}(\mathbf{V}))_j \|_q \Bigg\}.
\end{eqnarray}
Here $\mathcal{D}\subset\{1,2,\cdots,p\}$ contains prior information of hubs and we impose constraints that $\lambda_4 \leq \lambda_2,\lambda_5 \leq \lambda_3$. In the penalty, we give ``loose conditions'' for nodes in $\mathcal{D}$ so that they tend to be identified as hubs in our estimate. We will show that DHGL always performs better than HGL and keeps stable even when wrong prior information is provided.

\section{Methology}

\subsection{Discriminated Hub Graphical Lasso}

\subsubsection{Optimization Problem}

We continue the definitions and notations in Introduction. Combining Discriminated Hub Penalty Function with the loss function, we have the convex optimization problem
\begin{eqnarray}
\label{Eq:general}
 \underset{{\mathbf{\Theta}\in \mathcal{S}, \mathbf{V, Z}}} {\text{minimize}}
& & \Bigg\{ \ell(\mathbf{X},\mathbf{\Theta})  + \lambda_1 \| \mathbf{Z} - \text{diag}(\mathbf{Z})\|_1 +\lambda_2 \sum_{j\notin\mathcal{D}} \| (\mathbf{V}  - \text{diag}(\mathbf{V}))_j \|_1 \nonumber \\
&& + \lambda_3 \sum_{j\notin\mathcal{D}} \| (\mathbf{V}  - \text{diag}(\mathbf{V}))_j \|_q  + \lambda_4 \sum_{j\in\mathcal{D}} \| (\mathbf{V}  - \text{diag}(\mathbf{V}))_j \|_1 \nonumber \\
 &&  + \lambda_5 \nonumber \sum_{j\in\mathcal{D}} \| (\mathbf{V}  - \text{diag}(\mathbf{V}))_j \|_q \Bigg\} \;\;  \mbox{; } \text{subject to} \;\;  \mathbf{\Theta} = \mathbf{V}+\mathbf{V}^T+\mathbf{Z},
\end{eqnarray}
where $\mathcal{S}$ depends on the loss function $\ell(\mathbf{X},\mathbf{\Theta})$.

Similar to that in Tan et al. (2014), we encourage the solution of $\mathbf{Z}$ to be a sparse symmetric  matrix, and $\mathbf{V}$ to be a matrix with columns either entirely zero or almost entirely non-zero. Here $\lambda_1\geq 0$ controls the sparsity in $\mathbf{Z}$ \cite{Tan}. For variables in $\mathcal{D}$, $\lambda_5\geq 0$ controls the hub nodes selection, and $\lambda_4\geq 0$ controls the sparsity of each hub's connections to other nodes. Similar for $\lambda_2$ and $\lambda_3$ for variables not in $\mathcal{D}$. Here we set $q=2$, same as Tan et al. (2014) \cite{Tan,Yuan}.

As we can see, the ``discrimination'' involves using different tuning parameters controlling hub selection and hub sparsity for different columns in $\mathbf{V}$, or different variables (variables in $\mathcal{D}$ vs. not in $\mathcal{D}$). When $\mathcal{D}=\varnothing$, it reduces to the convex optimization problem corresponding to Hub Penalty Function in Tan et al. (2014). When $\lambda_2,\lambda_3,\lambda_4,\lambda_5\rightarrow\infty$, it reduces to the classical way to obtain a sparse graph estimate. Generally, $\mathcal{D}$ contains variables to be less penalized with $\lambda_4\leq\lambda_2$ and $\lambda_5\leq\lambda_3$.

 When $\mathbf{x}_1,\ldots,\mathbf{x}_n \stackrel{\small\mathrm{i.i.d.}}\sim N(\mathbf{0},\mathbf{\Sigma})$, we simply set $\ell(\mathbf{X},\mathbf{\Theta}) = -\log \det {\mathbf{\Theta}} + \mbox{trace}({\mathbf{S}} {\mathbf{\Theta}})$, where $\mathbf{S}$ is the empirical covariance matrix of $\mathbf{X}$. Then the Discriminated Hub Graphical Lasso (DHGL) optimization problem is
 \begin{eqnarray}
 \label{Eq:ggmhub}
  \underset{{\mathbf{\Theta}}\in \mathcal{S}} {\text{minimize}}
 & & \left \{ -\log \det  \mathbf{\Theta} + \text{trace}(\mathbf{S\Theta})  + \text{P}(\mathbf{\Theta}) \right\},
 \end{eqnarray}
 with $\mathcal{S} = \{\mathbf{\Theta}:\mathbf{\Theta} \succ 0 \text{ and } \mathbf{\Theta}=\mathbf{\Theta}^T \}$. Again, when $\mathcal{D}=\varnothing$, it reduces to Hub Graphical Lasso (HGL) in Tan et al. (2014), and when $\lambda_2,\lambda_3,\lambda_4,\lambda_5\rightarrow\infty$, it reduces to the classical Graphical Lasso (GL) \cite{Friedman}.

\subsubsection{Computational Complexity of ADMM}

 We can use Alternating Direction Method of Multipliers (ADMM) to solve the convex optimization problems proposed above \cite{Boyd}. The algorithm details as well as derivations are very similar to those of Algorithm 1 in Tan et al. (2014) \cite{Tan}. Details are displayed in Supplement Section 2.

 Also notice that the computational complexity is $\mathcal{O}(p^3)$ per iteration for solving DHGL, same in magnitude as HGL. We have also performed a simulation study to illustrate this. The simulation setup is very similar to that of ``DHGL applied with known hubs'' in Results and Analysis section except that we change $p$. In the first part, we set $p\in\{150,300,450\}$ and also change related parameters such as $n$, $r$ and $|\mathcal{H}|$ proportionally. In the second part, we set $p\in\{75,150,300\}$ but fix all other related parameters the same as the case when $p=150$. Results averaged over 25 replications are displayed in Table 1. We have found that the run time per iteration scale the similar way with $p$ regardless of other related parameters such as $n$. Also, the number of iterations increases as $p$ increase. Moreover, HGL and DHGL performs similarly in terms of run time and number of iterations.

\subsection{Tuning Parameter Selection}

 We can select tuning parameters $(\lambda_1,\lambda_2,\cdots,\lambda_5)$ by minimizing the same BIC-type quantity as proposed by Tan et al. (2014)
 \[
 \mathrm{BIC} (\hat{\mathbf{\Theta}},\hat{\mathbf{V}},\hat{\mathbf{Z}}) =-n \cdot \log \det (\hat{\mathbf{\Theta}}) + n \cdot \text{trace}(\mathbf{S}\hat{\mathbf{\Theta}}) + \log (n) \cdot |\hat{\mathbf{Z}}| + \log (n) \cdot \left(\nu+ c  \cdot [|\hat{\mathbf{V}}| - \nu ]\right),
 \]
 where the number of estimated hub nodes $\nu = \sum_{j=1}^p 1_{\{\|\hat{\mathbf{V}}_j\|_0 >1 \}}$, and $c\in(0,1)$ is a constant \cite{Tan}. Same as Tan et al. (2014), here we choose $c=0.2$ for our simulation experiments.

 In order for both optimal solutions for $\mathbf{V}$ and $\mathbf{Z}$, i.e., $\mathbf{V}^{\star}$ and $\mathbf{Z}^{\star}$ to be non-diagonal, we can select tuning parameters such that $$\frac{\lambda_2}{2}+\frac{\lambda_3}{2(p-1)^{\frac{1}{s}}}\leq\lambda_1\leq\frac{\lambda_2+\lambda_3}{2}, \lambda_4\leq\lambda_2, \lambda_5\leq\lambda_3, \dfrac{1}{s}+\dfrac{1}{q}=1.$$
 Detailed proof is displayed in Supplement Section 1.

\subsection{Measurement of Performance}

The precision matrix $\mathbf{\Theta}=\mathbf{\Sigma}^{-1}$ determines the conditional dependency between nodes \cite{Tan}. For $j\neq j'$, we know that if $|\mathbf{\Theta}_{jj'}|\neq 0$, a true edge between $j$ and $j'$ exists. If further $j\in\mathcal{H}$, this is also a hub edge. Then we set a tolerance value $t$ (in our experiment, we use $t=0.005$). If $|\hat{\mathbf{\Theta}}_{jj'}|>t$, an edge between $j$ and $j'$ is identified. The estimated hubs $\hat{\mathcal{H}}_r$ are defined by the set of nodes that have at least $r$ edges. Then we can define the number of correctly estimated edges, the proportion of correctly estimated hub edges, the proportion of correctly estimated hub nodes, the sum of squared errors between $\mathbf{\Theta}$ and $\hat{\mathbf{\Theta}}$ simply by their names, which are also nearly the same as those defined in Tan et al. (2014) \cite{Tan}. We have also defined the accuracy rate of hubs by the proportion of all $p$ nodes that are correctly estimated as hub nodes or correctly estimated as non-hub nodes.

\subsection{Obtain Set $\mathcal{D}$}

\subsubsection{DHGL Applied with Known Hubs}

Tan et al. (2014) propose HGL assuming they do not know a priori which nodes are hubs \cite{Tan}. However, before estimating the dependency structures among different nodes, we sometimes have domain knowledge of some dependency. In this case, we hope to use the prior information of hubs to estimate $\mathbf{\Theta}$ more accurately. In this section, we give an algorithm using DHGL to estimate $\mathbf{\Theta}$ when some hubs have been known.

\begin{algorithm}[htp]
\small
\caption{DHGL Applied with Known Hubs}

\begin{enumerate}
\item  Use HGL to get the estimated hubs $\hat{\mathcal{H}}_{\text{HGL}}$.
\item  Set $\mathcal{D}=\mathcal{K}\setminus \hat{\mathcal{H}}_{\text{HGL}}$, where $\mathcal{K}$ is a set of known hubs.
\item  If $\mathcal{D} \ne \varnothing$, use DHGL to estimate $\mathbf{\Theta}$ and get the estimated hubs $\hat{\mathcal{H}}_{\text{DHGL}}$, where $\lambda_1$, $\lambda_2$, $\lambda_3$ remain the same values as those in HGL and $\lambda_4$, $\lambda_5$ are selected using the BIC-type quantity. Then, set $\hat{\mathcal{H}}=\hat{\mathcal{H}}_{\text{HGL}} \bigcup \hat{\mathcal{H}}_{\text{DHGL}}$ as the set of estimated hubs. If $\mathcal{D}=\varnothing$, use the estimation in HGL directly.
\end{enumerate}
\end{algorithm}

In HGL, when $\lambda_2$ and $\lambda_3$ become smaller, it is more likely to find more edges and hubs since all of the entries in $\hat{\mathbf{\Theta}}_{\text{HGL}}$ have less penalty and tend to become larger. However, while we find more correct edges and hubs as $\lambda_2$ and $\lambda_3$ become smaller, the number of wrong edges and hubs we find also increase. Thus, if we have known some true hubs or some prior information of nodes which are likely to be hubs, we hope to give ``loose conditions'' only for the entries of their columns of $\hat{\mathbf{\Theta}}$ to become larger rather than all of the nodes. That's why we let $\mathcal{D}$ contain the prior information and propose $\lambda_4$ and $\lambda_5$ which are not greater than $\lambda_2$ and $\lambda_3$ respectively. If a known hub does not belong to $\hat{\mathcal{H}}_{\text{HGL}}$ which means the corresponding column of $\hat{\mathbf{\Theta}}_{\text{HGL}}$ contains many zeros, some entries of the corresponding column of $\hat{\mathbf{\Theta}}_{\text{DHGL}}$ will become larger and more correct edges will be identified. On the other hand, if a node belongs to $\hat{\mathcal{H}}_{\text{HGL}}$ which means the corresponding column of $\hat{\mathbf{\Theta}}_{\text{HGL}}$ contains many nonzeros and we still set it in $\mathcal{D}$, the corresponding column of $\hat{\mathbf{\Theta}}_{\text{DHGL}}$ will be overamplified which may even become worse than $\hat{\mathbf{\Theta}}_{\text{HGL}}$. Thus, we set $\mathcal{D}=\mathcal{K}\setminus \hat{\mathcal{H}}_{\text{HGL}}$ to avoid this problem. Furthermore, while the algorithm finds more correct edges of the discriminated nodes, a few correct edges found by HGL may disappear so that some nodes in $\hat{\mathcal{H}}_{\text{HGL}}$ may not belong to $\hat{\mathcal{H}}_{\text{DHGL}}$. Thus, we need to combine $\hat{\mathcal{H}}_{\text{HGL}}$ and $\hat{\mathcal{H}}_{\text{DHGL}}$ so that we will not lose useful information of hubs.

\subsubsection{DHGL Applied without Known Hubs}

In the previous section, we discuss the application of DHGL when some hubs are known. However, we often do not have any prior information of hubs. In this section, we give an algorithm using GL to provide prior information and DHGL to estimate $\mathbf{\Theta}$.

\begin{algorithm}[htp]
\small
\caption{DHGL Applied without Known Hubs}

\begin{enumerate}
\item  Use HGL to get the estimated hubs $\hat{\mathcal{H}}_{\text{HGL}}$.
\item  (Prior Information Screening) Adjust regularization parameter $\lambda$ of GL from large to small until $|\hat{\mathcal{H}}_{\text{GL},\lambda}\setminus \hat{\mathcal{H}}_{\text{HGL}}|>0$ and
$|\hat{\mathcal{H}}_{\text{GL},\lambda}\bigcup \hat{\mathcal{H}}_{\text{HGL}}| \leq\max\{|\hat{\mathcal{H}}_{\text{HGL}}|+a,b|\hat{\mathcal{H}}_{\text{HGL}}|\}$
where $a\in\boldsymbol{N}_+$, $b>1$ but $b\approx1$ and $\hat{\mathcal{H}}_{\text{GL},\lambda}$ is the set of estimated hubs by GL with the parameter $\lambda$.
\item  Set $\mathcal{D}=\hat{\mathcal{H}}_{\text{GL},\lambda}\setminus \hat{\mathcal{H}}_{\text{HGL}}$ which is non-empty.
\item  Use DHGL to estimate $\mathbf{\Theta}$, where $\lambda_1$, $\lambda_2$, $\lambda_3$, $\lambda_4$ remain the same values as those in HGL and $\lambda_5$ is selected using the BIC-type quantity.
\end{enumerate}
\end{algorithm}

The BIC-type quantity can help us select the tuning parameters to get a relatively accurate estimate of $\mathbf{\Theta}$. However, $|\hat{\mathcal{H}}_{\text{HGL}}|$ is always not more than the number of true hubs $\lvert \mathcal{H} \rvert$ because of the penalty term of the number of hubs in the BIC-type quantity, especially when $\lvert \mathcal{H} \rvert$ is relatively large. Based on this fact, we try to find extra prior information of hubs which are not in $\hat{\mathcal{H}}_{\text{HGL}}$. Since GL is fast (less than $\mathcal{O}(p^3)$ for sparse $\mathbf{\Theta}$) \cite{Friedman}, we can adjust the regularization parameter of GL from large to small quickly until we get nodes that belong to $\hat{\mathcal{H}}_{\text{GL},\lambda}$ rather than $\hat{\mathcal{H}}_{\text{HGL}}$. These extra nodes are likely to be true hubs if $|\hat{\mathcal{H}}_{\text{HGL}}|<|\mathcal{H}|$ although there is no guarantee for this. Thus, we set the extra nodes in $\mathcal{D}$. Since the prior information provided by GL may be wrong, we keep $\lambda_4=\lambda_2$ for conservatism and we only need to select $\lambda_5$ using the BIC-type quantity in DHGL. If the extra nodes are true hubs, $\lambda_5$ tends to be small to make the BIC-type quantity small so that more true edges and hubs will be found. If the extra nodes are not true hubs, $\lambda_5$ tends to keep the same as $\lambda_3$ and the estimation of $\mathbf{\Theta}$ will be the same as that in HGL. In a word, the new algorithm tends to accept correct prior information to make improvement and reject false one to keep stable.

\section{Results and Analysis}

In the simulation studies, the precision matrix $\mathbf{\Theta}$ and the set of indices of hub nodes $\mathcal{H}$ are given, and we estimate $\hat{\mathbf{\Theta}}$ using our approaches displayed above.

\subsection{DHGL Applied with Known Hubs}

In the experiment, firstly, we use the simulation set-up I in Tan et al. (2014) with $p=150$, $n=50$, $r=30$, $\lvert \mathcal{H} \rvert=5$ and suppose we know two true hubs randomly \cite{Tan}. We fix $\lambda_1=0.4$ and consider two cases where $\lambda_3=1$ and $\lambda_3=1.5$. In each case, $\lambda_2$ ranges from 0.1 to 0.7. In DHGL, for simplicity of the experiment, we do not use the BIC-type quantity to select $\lambda_4$ and $\lambda_5$ but set $\lambda_4$ relatively smaller than $\lambda_2$ and $\lambda_5=0.1$. For every set of tuning parameters, we conduct 50 simulations and average the results. Figure 1 shows the comparison of measures of performance between HGL and DHGL.

From Figure 1, we see that for every set of tuning parameters, the number of correctly estimated edges, the proportion of correctly estimated hub edges and the sum of squared errors perform better in DHGL than those in HGL. For the effective proportion of correctly estimated hubs and the effective accuracy rate of hubs, HGL performs better than DHGL reasonably since the discriminated nodes (known hubs) are not considered in the calculation of the two measures. In order not to lose useful information, we combine the hubs found by HGL and DHGL so that we can always get better information of hubs using our method.

Moreover, we consider three other cases for $(p,n)\in \{(200,50),(200,100),(300,100)\}$, where $\lambda_2=0.4$, $\lambda_3=1$, $\lambda_4=0.2$, $r=\frac{p}{5}$ and other parameters are set to the same. Figure 2 displays performances of 50 simulations when $(p,n)=(300,100)$. All of the three results are similar to the case when $(p,n)=(150,50)$.

\subsection{DHGL Applied without Known Hubs}

In the experiment, we use the simulation set-up I in Tan et al. (2014) with $p=150$, $n=50$, $r=30$ and $\lvert \mathcal{H} \rvert \in \{5,10\}$ \cite{Tan}. We fix $\lambda_1=0.4$, $\lambda_3=1$ and select $\lambda_2$ from $[0.05,0.15]$ using the BIC-type quantity. Then for prior information screening, we set $a=2$ and $b=1.1$. In DHGL, we select $\lambda_5$ from $[0.5,1]$ using the BIC-type quantity with $c=0.1$. Figure 3 and 4 display the comparison of 50 simulations between HGL and DHGL when $\lvert \mathcal{H} \rvert=10$ and $5$ respectively.

From Figure 3, we see that for $\lvert \mathcal{H} \rvert=10$ which is relatively large, all of the five measures in DHGL outperform those in HGL. Even when the accuracy rate of hubs in HGL equal to 1 (4 of 50) which means the discriminated nodes are not true hubs, all of the performances in DHGL do not degenerate.

From Figure 4, we see that for $\lvert \mathcal{H}\rvert=5$ which is relatively small, the majority of accuracy rates of hubs in HGL equal to 1 (31 of 50). Among these cases, only 5 accuracy rates of hubs in DHGL degenerate and the other measures of the 5 also degenerate slightly. However, when the accuracy rates of hubs in HGL is less than 1 which means $\lvert \hat{\mathcal{H}}_{\text{HGL}} \rvert < \lvert \mathcal{H} \rvert$, all the measures of the majority (14 of 19) outperform those in HGL obviously.

These two results display both the advantages and robustness of DHGL applied without known hubs. When $\lvert \hat{\mathcal{H}}_{\text{HGL}} \rvert <\lvert \mathcal{H}\rvert$, DHGL can always perform better than HGL. On the other hand, when $\lvert \hat{\mathcal{H}}_{\text{HGL}} \rvert =\lvert \mathcal{H}\rvert$ which is not always the case, the performance of DHGL seldom degenerates and always keep the same as that of HGL.

\section{Discussion}

Based on Hub Graphical Lasso proposed by Tan et al. (2014), we propose Discriminated Hub Graphical Lasso to estimate the precision matrix when some prior information of hubs is known. For the two situations where some hubs are known or no hubs are known, we propose two algorithms respectively to make some improvement in the measures of performance given by Tan et al. (2014) and us. The improvement essentially results from the increase of the number of tuning parameters for the prior information and the BIC-type quantity proposed by Tan et al. (2014) will be smaller than that in HGL by selecting the optimal parameters. The computational complexity is still $\mathcal{O}(p^3)$ per iteration in DHGL, same as that in HGL. However, the two algorithms need us to use both HGL and DHGL which will certainly cost more time but in the same scale. Moreover, in both algorithms, only the prior information of hubs which have not been found by HGL are set in $\mathcal{D}$ so sometimes we cannot make use of the prior information when $\mathcal{D}=\varnothing$. Thus, it remains open problems how to use DHGL only to make improvement and how to use all prior information of hubs.

For Algorithm 2, we do not give a certain method about how to select the optimal a and b when screening prior information using Graphical Lasso. Instead, we select relatively small $a$ and $b$ for conservatism which will limit the performance of the algorithm. For the future work, we hope to figure out how to select the optimal $a$ and $b$, which means the difference between the number of estimated hubs in HGL and the number of true hubs should be studied more deeply. Moreover, we will try to find other methods instead of GL to search for prior information of hubs.

\section*{Tables and Figures}

\begin{figure}[h]
\begin{center}
\includegraphics[width=1.0\linewidth]{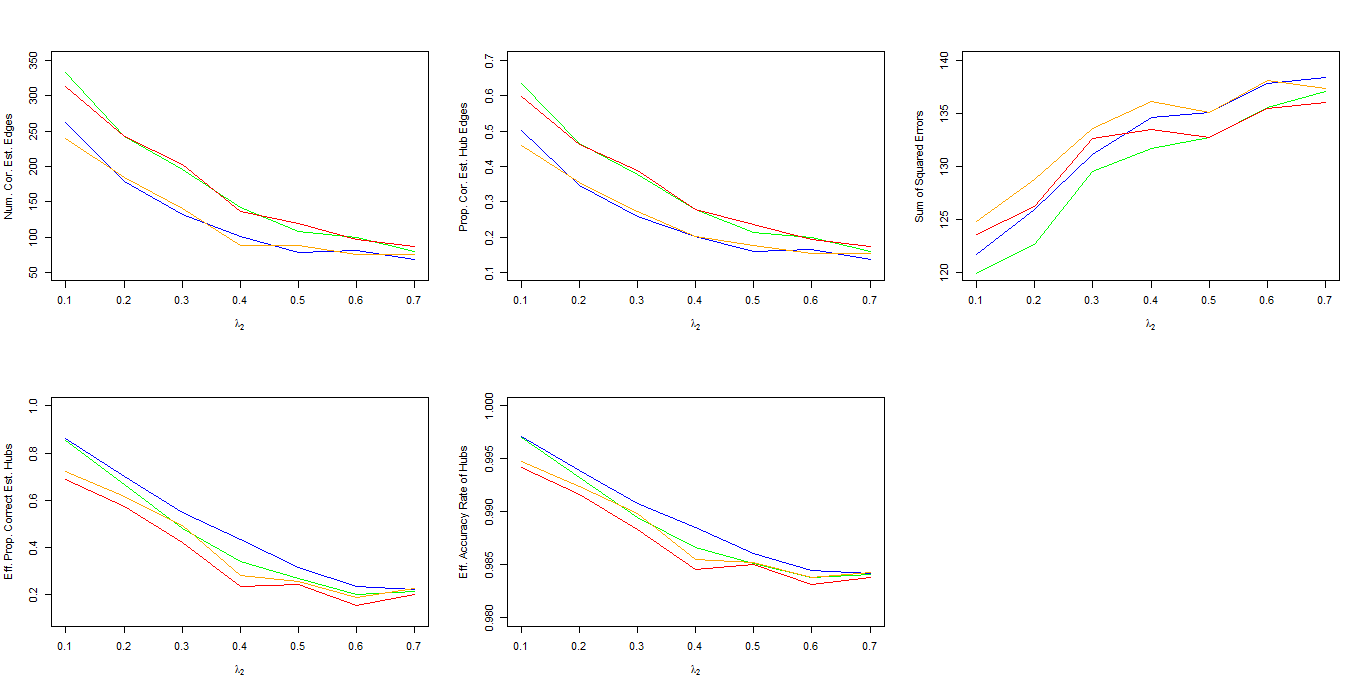}
\end{center}
\caption{Measures of performance (Number of correctly estimated edges, Proportion of correctly estimated hub edges, Effective proportion of correctly estimated hub nodes, Sum of squared errors between $\mathbf{\Theta}$ and $\hat{\mathbf{\Theta}}$, Effective accuracy rate of hubs) of HGL and DHGL when some hubs are known, where Effective proportion of correctly estimated hub nodes and Effective accuracy rate of hubs are calculated not considering known hubs. The $x$-axes display $\lambda_2$ and the $y$-axes display the five measures. The blue, green, orange, red lines correspond to $(\lambda_3=1, \text{HGL}), (\lambda_3=1, \text{DHGL}), (\lambda_3=1.5, \text{HGL}), (\lambda_3=1.5, \text{DHGL})$ respectively. 50 simulations are conducted for each set of parameters.}
\end{figure}

\begin{figure}[h]
\begin{center}
\includegraphics[width=1.0\linewidth]{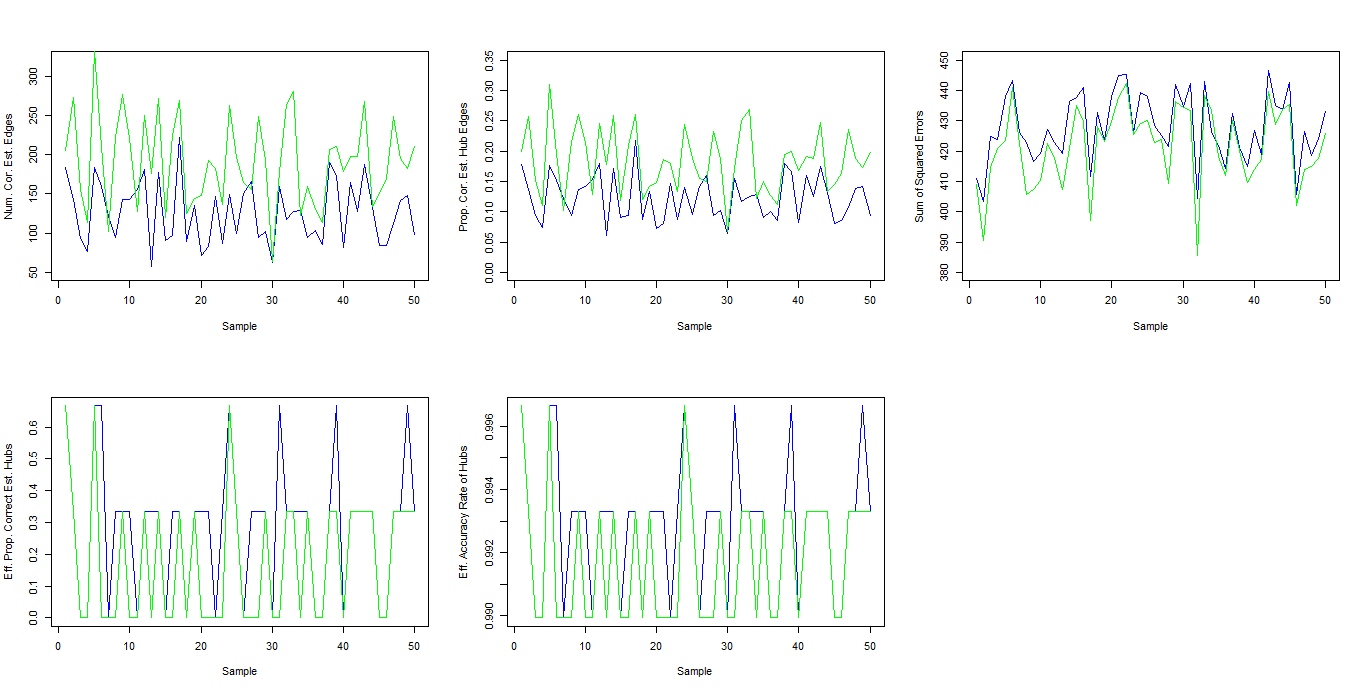}
\end{center}
\caption{Measures of performance (Number of correctly estimated edges, Proportion of correctly estimated hub edges, Effective proportion of correctly estimated hub nodes, Sum of squared errors between $\mathbf{\Theta}$ and $\hat{\mathbf{\Theta}}$, Effective accuracy rate of hubs) of HGL and DHGL when some hubs are known, where Effective proportion of correctly estimated hub nodes and Effective accuracy rate of hubs are calculated not considering known hubs. The $x$-axes display the 50 simulations and the $y$-axes display the five measures. The results are for p=300, n=100 and $(\lambda_1,\lambda_2,\lambda_3,\lambda_4,\lambda_5)=(0.4,0.4,1,0.2,0.1)$. The blue and green lines correspond to HGL and DHGL respectively.}
\end{figure}

\begin{figure}[h]
\begin{center}
\includegraphics[width=1.0\linewidth]{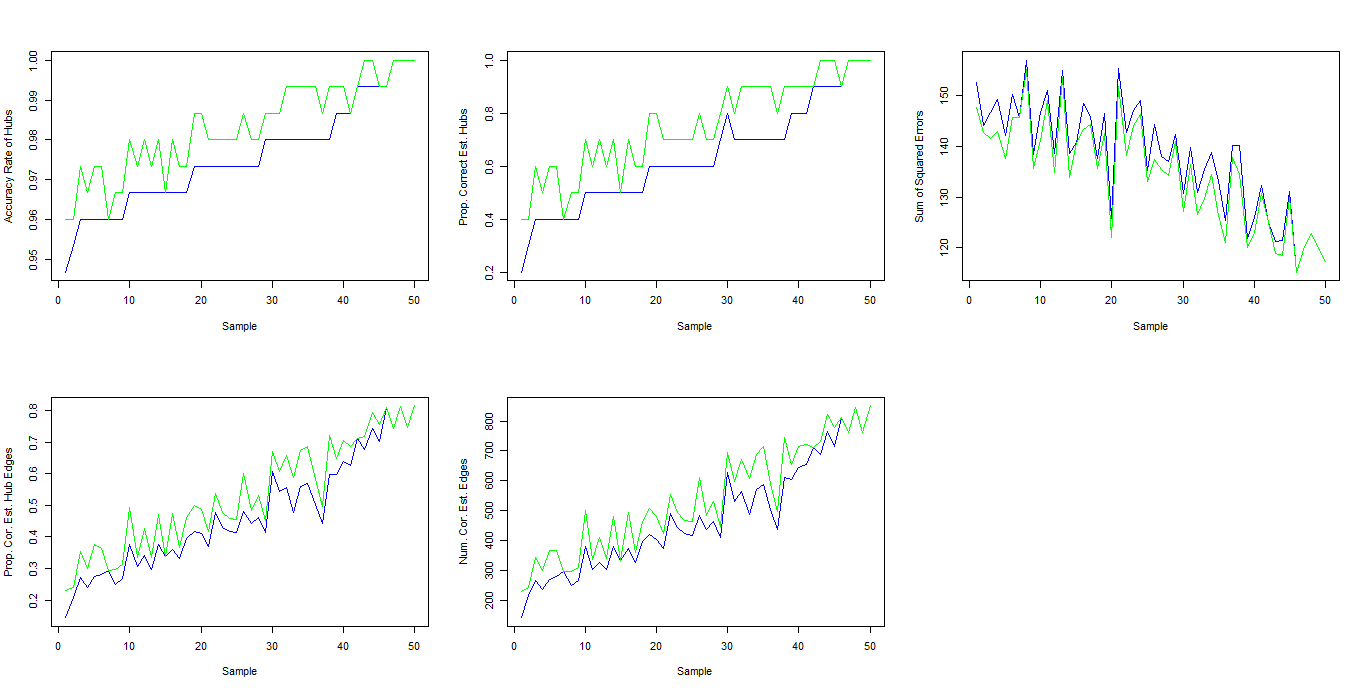}
\end{center}
\caption{Measures of performance (Number of correctly estimated edges, Proportion of correctly estimated hub edges, Proportion of correctly estimated hub nodes, Sum of squared errors between $\mathbf{\Theta}$ and $\hat{\mathbf{\Theta}}$, Accuracy rate of hubs) of HGL and DHGL when no hubs are known. The $x$-axes display the 50 simulations and the $y$-axes display the five measures. The results are for p=150, n=50 and $|\mathcal{H}|=10$. The blue and green lines correspond to HGL and DHGL respectively. Here, we sort the accuracy rate of hubs of the 50 samples.}
\end{figure}

\begin{figure}[h]
\begin{center}
\includegraphics[width=1.0\linewidth]{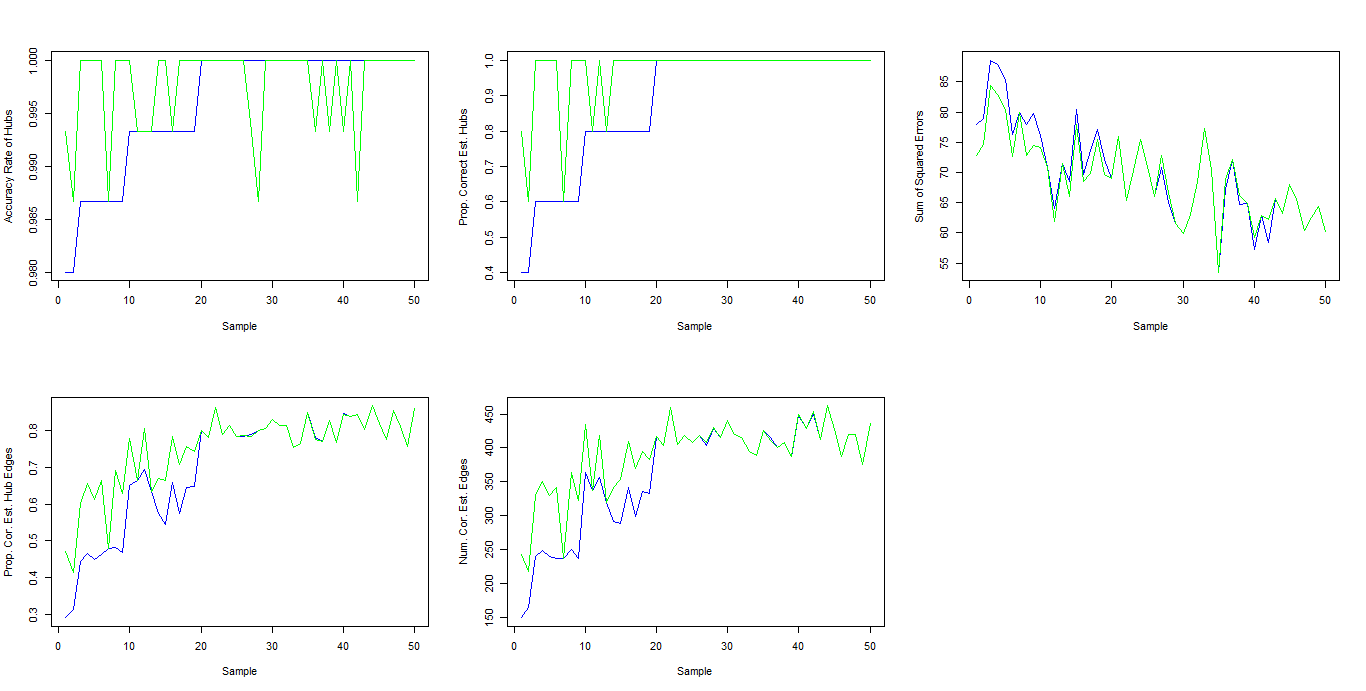}
\end{center}
\caption{Measures of performance (Number of correctly estimated edges, Proportion of correctly estimated hub edges, Proportion of correctly estimated hub nodes, Sum of squared errors between $\mathbf{\Theta}$ and $\hat{\mathbf{\Theta}}$, Accuracy rate of hubs) of HGL and DHGL when no hubs are known. The $x$-axes display the 50 simulations and the $y$-axes display the five measures. The results are for p=150, n=50 and $|\mathcal{H}|=5$. The blue and green lines correspond to HGL and DHGL respectively. Here, we sort the accuracy rate of hubs of the 50 samples.}
\end{figure}

\begin{table}[t]
\caption{Simulation study on run time and number of iterations of HGL and DHGL. The simulation setup is very similar to that of ``DHGL applied with known hubs'' in Analysis and Results section except that we change $p$. Runs represents the times that DHGL is used among 25 simulations.}
\label{sample-table}
\begin{center}
\begin{tabular}{cc|cc}
\hline  &  & \multicolumn{2}{c}{(Run time, Iteration, Time per iteration, Runs)} \\ $p$ & $(n,r,|\mathcal{H}|)$ & HGL & DHGL
\\ \hline
150 & $(50,30,5)$ & $(1.065,42.7,0.025,25)$ & $(1.093,43.6,0.025,23)$ \\
300 & $(100,60,10)$ & $(6.139,56.6,0.108,25)$ & $(6.281,57.4,0.109,25)$ \\ 450 & $(150,90,15)$ & $(19.46,67.9,0.287,25)$ & $(19.62,68.3,0.287,25)$ \\ \hline 75 & $(50,30,5)$ & $(0.131,33.2,0.004,25)$ & $(0.136,34.0,0.004,25)$ \\ 150 & $(50,30,5)$ & $(0.756,43.6,0.017,25)$ & $(0.771,44.1,0.018,24)$ \\ 300 & $(50,30,5)$ & $(6.700,58.1,0.115,25)$ & $(6.767,58.3,0.116,23)$\\\hline
\end{tabular}
\end{center}
\end{table}

\section{Supplement}

\subsection{Proof of Theorem in Selecting Tunning Parameters}

Recall that the optimization problem is
\begin{eqnarray}
\label{Eq:general}
 \underset{{\mathbf{\Theta}\in \mathcal{S}, \mathbf{V, Z}}} {\text{minimize}}
& & \Bigg\{ \ell(\mathbf{X},\mathbf{\Theta})  + \lambda_1 \| \mathbf{Z} - \text{diag}(\mathbf{Z})\|_1 +\lambda_2 \sum_{j\notin\mathcal{D}} \| (\mathbf{V}  - \text{diag}(\mathbf{V}))_j \|_1  \nonumber \\
&& + \lambda_3 \sum_{j\notin\mathcal{D}} \| (\mathbf{V}  - \text{diag}(\mathbf{V}))_j \|_q + \lambda_4 \sum_{j\in\mathcal{D}} \| (\mathbf{V}  - \text{diag}(\mathbf{V}))_j \|_1\nonumber  \\&& + \lambda_5 \sum_{j\in\mathcal{D}} \| (\mathbf{V}  - \text{diag}(\mathbf{V}))_j \|_q \Bigg\} \;\; \nonumber \\
 \text{subject to} \;\; &&   \mathbf{\Theta} = \mathbf{V}+\mathbf{V}^T+\mathbf{Z},
\end{eqnarray}
where the set $\mathcal{S}$ depends on the loss function $\ell(\mathbf{X},\mathbf{\Theta})$.

Let $\mathbf{Z}^{\star}$ and $\mathbf{V}^{\star}$ denote the optimal solution for $\mathbf{Z}$ and $\mathbf{V}$ in (\ref{Eq:general}) and $\frac{1}{s}+\frac{1}{q}=1$.

{\bf Theorem 1:} If $\lambda_{1}<\frac{\lambda_2}{2}+\frac{\lambda_3}{2(p-1)^{1/s}}$, then $\mathbf{V}_{ij}^{\star}=0$, for $\forall j\notin \mathcal{D}, \forall i\neq j$.

{\bf Proof:} Let $(\mathbf{\Theta}^{\star},\mathbf{Z}^{\star},\mathbf{V}^{\star})$ be the solution to (\ref{Eq:general}) and suppose $\exists j\notin \mathcal{D},i\neq j$ such that $\mathbf{V}_{ij}^{\star}\neq 0$.

Let $\hat{\mathbf{V}}_{ij}=\mathbf{V}_{ij}^{\star}$ for $\forall j\in \mathcal{D}, \forall i\neq j$ and $\hat{\mathbf{V}}_{ij}=0$ for $\forall j\notin \mathcal{D}, \forall i\neq j$ and $\text{diag}(\hat{\mathbf{V}})=\text{diag}(\mathbf{V}^{\star}).$

Also we construct $\hat{\mathbf{Z}}$ as follows\\
\begin{equation}
\hat{\mathbf{Z}}_{ij}=
  \begin{cases}
  \mathbf{Z}_{ij}^{\star}+\mathbf{V}_{ij}^{\star}+\mathbf{V}_{ji}^{\star} &\mbox{for $\forall j\notin \mathcal{D}, \forall i\notin \mathcal{D}, i\neq j$}\\
  \mathbf{Z}_{ij}^{\star} &\mbox{otherwise}
  \end{cases}
\end{equation}

Then we have $\mathbf{\Theta}^{\star}=\hat{\mathbf{V}}+\hat{\mathbf{V}}^{T}+\hat{\mathbf{Z}}$. We now show that $(\mathbf{\Theta}^{\star},\hat{\mathbf{Z}},\hat{\mathbf{V}})$ has a smaller objective value than $(\mathbf{\Theta}^{\star},\mathbf{Z}^{\star},\mathbf{V}^{\star})$, which is a contradiction.

Firstly, we have\\
\begin{equation*}
\begin{split}
&\lambda_{1} \left\|\hat{\mathbf{Z}}-\text{diag}(\hat{\mathbf{Z}}) \right\|_{1}+\lambda_{2}\sum_{j\notin \mathcal{D}}\left\|(\hat{\mathbf{V}}-\text{diag}(\hat{\mathbf{V}}))_{j} \right\|_1
+ \lambda_{4}\sum_{j\in \mathcal{D}}\left\|(\hat{\mathbf{V}}-\text{diag}(\hat{\mathbf{V}}))_{j} \right\|_1\\+ &\lambda_{3}\sum_{j\notin \mathcal{D}}\left\|(\hat{\mathbf{V}}-\text{diag}(\hat{\mathbf{V}}))_{j} \right\|_{q}
+\lambda_{5}\sum_{j\in \mathcal{D}}\left\|(\hat{\mathbf{V}}-\text{diag}(\hat{\mathbf{V}}))_{j} \right\|_{q}\\
=&\lambda_{1} \left\|\hat{\mathbf{Z}}-\text{diag}(\hat{\mathbf{Z}}) \right\|_{1}+0
+ \lambda_{4}\sum_{j\in \mathcal{D}}\left\|(\mathbf{V}^{\star}-\text{diag}(\mathbf{V}^{\star}))_{j} \right\|_1+ 0
+\lambda_{5}\sum_{j\in \mathcal{D}}\left\|(\mathbf{V}^{\star}-\text{diag}(\mathbf{V}^{\star}))_{j} \right\|_{q}\\
\leq &\lambda_{1} \left\|\mathbf{Z}^{\star}-\text{diag}(\mathbf{Z}^{\star}) \right\|_{1}+2 \lambda_1\sum_{i \notin \mathcal{D}, j \notin \mathcal{D}, i\neq j} | \mathbf{V}_{ij}^{\star} |
+ \lambda_{4}\sum_{j\in \mathcal{D}}\left\|(\mathbf{V}^{\star}-\text{diag}(\mathbf{V}^{\star}))_{j} \right\|_1\\
+&\lambda_{5}\sum_{j\in \mathcal{D}}\left\|(\mathbf{V}^{\star}-\text{diag}(\mathbf{V}^{\star}))_{j} \right\|_{q}\\
\leq &\lambda_{1} \left\|\mathbf{Z}^{\star}-\text{diag}(\mathbf{Z}^{\star}) \right\|_{1}+2 \lambda_1 \sum_{j\notin \mathcal{D}} \left\|(\mathbf{V}^{\star}-\text{diag}(\mathbf{V}^{\star}))_j \right\|_{1}
+ \lambda_{4}\sum_{j\in \mathcal{D}}\left\|(\mathbf{V}^{\star}-\text{diag}(\mathbf{V}^{\star}))_{j} \right\|_1\\
+ &\lambda_{5}\sum_{j\in \mathcal{D}}\left\|(\mathbf{V}^{\star}-\text{diag}(\mathbf{V}^{\star}))_{j} \right\|_{q}\\
=&\lambda_{1} \left\|\mathbf{Z}^{\star}-\text{diag}(\mathbf{Z}^{\star}) \right\|_{1}+\lambda_2 \sum_{j\notin \mathcal{D}} \left\|(\mathbf{V}^{\star}-\text{diag}(\mathbf{V}^{\star}))_j \right\|_{1}+(2 \lambda_1-\lambda_2)\sum_{j\notin \mathcal{D}} \left\|(\mathbf{V}^{\star}-\text{diag}(\mathbf{V}^{\star}))_j \right\|_{1}\\
+ &\lambda_{4}\sum_{j\in \mathcal{D}}\left\|(\mathbf{V}^{\star}-\text{diag}(\mathbf{V}^{\star}))_{j} \right\|_1+\lambda_{5}\sum_{j\in \mathcal{D}}\left\|(\mathbf{V}^{\star}-\text{diag}(\mathbf{V}^{\star}))_{j} \right\|_{q}.\\
\end{split}
\end{equation*}

By Holder's Inequality $\mathbf{x}^{T}\mathbf{y} \leq \left\|\mathbf{x} \right\|_q \left\|\mathbf{y} \right\|_s$, where $\frac{1}{s}+\frac{1}{q}=1$ and $ \mathbf{x},\mathbf{y} \in \mathbb{R}^{p-1}$. Let $\mathbf{y}=\text{sign}(\mathbf{x})$, then we have$\left\|\mathbf{x} \right\|_1\leq (p-1)^{\frac{1}{s}}\left\|\mathbf{x} \right\|_q.$

Hence, $\sum_{j\notin \mathcal{D}}\left\|(\mathbf{V}^{\star}-\text{diag}(\mathbf{V}^{\star}))_{j} \right\|_{1}\leq (p-1)^{\frac{1}{s}}\sum_{j\notin \mathcal{D}}\left\|(\mathbf{V}^{\star}-\text{diag}(\mathbf{V}^{\star}))_{j} \right\|_{q}$.

Hence,
\begin{equation*}
\begin{split}
&\lambda_{1} \left\|\hat{\mathbf{Z}}-\text{diag}(\hat{\mathbf{Z}}) \right\|_{1}+\lambda_{2}\sum_{j\notin \mathcal{D}}\left\|(\hat{\mathbf{V}}-\text{diag}(\hat{\mathbf{V}}))_{j} \right\|_1
+ \lambda_{4}\sum_{j\in \mathcal{D}}\left\|(\hat{\mathbf{V}}-\text{diag}(\hat{\mathbf{V}}))_{j} \right\|_1\\+ &\lambda_{3}\sum_{j\notin \mathcal{D}}\left\|(\hat{\mathbf{V}}-\text{diag}(\hat{\mathbf{V}}))_{j} \right\|_{q}
+\lambda_{5}\sum_{j\in \mathcal{D}}\left\|(\hat{\mathbf{V}}-\text{diag}(\hat{\mathbf{V}}))_{j} \right\|_{q}\\
\leq &\lambda_{1} \left\|\mathbf{Z}^{\star}-\text{diag}(\mathbf{Z}^{\star}) \right\|_{1}+\lambda_{2}\sum_{j\notin \mathcal{D}}\left\|(\mathbf{V}^{\star}-\text{diag}(\mathbf{V}^{\star}))_{j} \right\|_1
+ \lambda_{4}\sum_{j\in \mathcal{D}}\left\|(\mathbf{V}^{\star}-\text{diag}(\mathbf{V}^{\star}))_{j} \right\|_1\\+ &(2\lambda_1-\lambda_2)(p-1)^{\frac{1}{s}}\sum_{j\notin \mathcal{D}}\left\|(\mathbf{V}^{\star}-\text{diag}(\mathbf{V}^{\star}))_{j} \right\|_{q}
+\lambda_{5}\sum_{j\in \mathcal{D}}\left\|(\mathbf{V}^{\star}-\text{diag}(\mathbf{V}^{\star}))_{j} \right\|_{q}\\
< &\lambda_{1} \left\|\mathbf{Z}^{\star}-\text{diag}(\mathbf{Z}^{\star}) \right\|_{1}+\lambda_{2}\sum_{j\notin \mathcal{D}}\left\|(\mathbf{V}^{\star}-\text{diag}(\mathbf{V}^{\star}))_{j} \right\|_1
+ \lambda_{4}\sum_{j\in \mathcal{D}}\left\|(\mathbf{V}^{\star}-\text{diag}(\mathbf{V}^{\star}))_{j} \right\|_1\\+ &\lambda_3\sum_{j\notin \mathcal{D}}\left\|(\mathbf{V}^{\star}-\text{diag}(\mathbf{V}^{\star}))_{j} \right\|_{q}
+\lambda_{5}\sum_{j\in \mathcal{D}}\left\|(\mathbf{V}^{\star}-\text{diag}(\mathbf{V}^{\star}))_{j} \right\|_{q}.\\
\end{split}
\end{equation*}

For the first inequality, we assume $2\lambda_1 \ge \lambda_2$. On the other hand, if $2\lambda_1 < \lambda_2$, the last inequality holds obviously. Thus, this leads to a contradiction.

{\bf Theorem 2:} If $\lambda_1>\frac{\lambda_2+\lambda_3}{2}, \lambda_4\leq \lambda_2, \lambda_5\leq \lambda_3$, then $\mathbf{Z}^{\star}$ is a diagonal matrix.

{\bf Proof:} Let $(\mathbf{\Theta}^{\star},\mathbf{Z}^{\star},\mathbf{V}^{\star})$ be the solution to (\ref{Eq:general}) and suppose  $\mathbf{Z}^{\star}$ is not a diagonal matrix. Note that $\mathbf{Z}^{\star}$ is symmetric since
$\mathbf{\Theta}\in \mathcal{S}, \mathcal{S}=\{\mathbf{\Theta}: \mathbf{\Theta}\succ0, \mathbf{\Theta}=\mathbf{\Theta}^{T}\}$. Let $\hat{\mathbf{Z}}=\text{diag}(\mathbf{Z}^{\star})$ and also construct $\hat{\mathbf{V}}$ as follows
\begin{equation}
\hat{\mathbf{V}}_{ij}=
\begin{cases}
\mathbf{V}^{\star}_{ij}+\frac{\mathbf{Z}^{\star}_{ij}}{2} &\mbox{$i\neq j$}\\
\mathbf{V}^{\star}_{jj} &\mbox{otherwise}
\end{cases}
\end{equation}

Then we have $\mathbf{\Theta}^{\star}=\hat{\mathbf{Z}}+\hat{\mathbf{V}}+\hat{\mathbf{V}}^{T}$. We now show that $(\mathbf{\Theta}^{\star},\hat{\mathbf{Z}},\hat{\mathbf{V}})$ has a smaller objective value than $(\mathbf{\Theta}^{\star},\mathbf{Z}^{\star},\mathbf{V}^{\star})$ which is a contradiction.

Note that
\begin{equation*}
\begin{split}
&\lambda_{1} \left\|\hat{\mathbf{Z}}-\text{diag}(\hat{\mathbf{Z}}) \right\|_{1}+\lambda_2 \sum_{j\notin \mathcal{D}}\left\|(\hat{\mathbf{V}}-\text{diag}(\hat{\mathbf{V}}))_j \right\|_{1}
+ \lambda_{4}\sum_{j\in \mathcal{D}}\left\|(\hat{\mathbf{V}}-\text{diag}(\hat{\mathbf{V}}))_{j} \right\|_1\\
=&\lambda_2 \sum_{j\notin \mathcal{D}}\left\|(\hat{\mathbf{V}}-\text{diag}(\hat{\mathbf{V}}))_j \right\|_{1}
+ \lambda_{4}\sum_{j\in \mathcal{D}}\left\|(\hat{\mathbf{V}}-\text{diag}(\hat{\mathbf{V}}))_{j} \right\|_1\\
=&\lambda_2 \sum_{j\notin \mathcal{D}, i\neq j}|(\mathbf{V}^{\star}_{ij}+\frac{\mathbf{Z}^{\star}_{ij}}{2}) |
+ \lambda_{4}\sum_{j\in \mathcal{D}, i\neq j}|(\mathbf{V}^{\star}_{ij}+\frac{\mathbf{Z}^{\star}_{ij}}{2})|\\
\leq &\lambda_2\sum_{j\notin \mathcal{D}}\left\|(\mathbf{V}^{\star}-\text{diag}(\mathbf{V}^{\star}))_j \right\|_{1}+
\lambda_{4}\sum_{j\in \mathcal{D}}\left\|(\mathbf{V}^{\star}-\text{diag}(\mathbf{V}^{\star}))_{j} \right\|_1
+\frac{\lambda_2}{2}\sum_{j\notin \mathcal{D}, i\neq j}|\mathbf{Z}_{ij}^{\star}|+\frac{\lambda_4}{2}\sum_{j \in \mathcal{D},i\neq j}|\mathbf{Z}_{ij}^{\star}|\\
\leq &\lambda_2\sum_{j\notin \mathcal{D}}\left\|(\mathbf{V}^{\star}-\text{diag}(\mathbf{V}^{\star}))_j \right\|_{1}+
\lambda_{4}\sum_{j\in \mathcal{D}}\left\|(\mathbf{V}^{\star}-\text{diag}(\mathbf{V}^{\star}))_{j} \right\|_1
+\frac{\lambda_2}{2}\left\|(\mathbf{Z}^{\star}-\text{diag}(\mathbf{Z}^{\star})) \right\|_1.
\end{split}
\end{equation*}

And also,
\begin{equation*}
\begin{split}
&\lambda_{3}\sum_{j\notin \mathcal{D}}\left\|(\hat{\mathbf{V}}-\text{diag}(\hat{\mathbf{V}}))_{j} \right\|_{q}
+\lambda_{5}\sum_{j\in \mathcal{D}}\left\|(\hat{\mathbf{V}}-\text{diag}(\hat{\mathbf{V}}))_{j} \right\|_{q}\\
\leq &\lambda_{3}\sum_{j\notin \mathcal{D}}\left\|(\mathbf{V}^{\star}-\text{diag}(\mathbf{V}^{\star}))_{j} \right\|_{q}
+\frac{\lambda_3}{2}\sum_{j\notin \mathcal{D}}\left\|(\mathbf{Z}^{\star}-\text{diag}(\mathbf{Z}^{\star}))_{j} \right\|_{q}\\
+&\lambda_{5}\sum_{j\in \mathcal{D}}\left\|(\mathbf{V}^{\star}-\text{diag}(\mathbf{V}^{\star}))_{j} \right\|_{q}
+\frac{\lambda_5}{2}\sum_{j\in \mathcal{D}}\left\|(\mathbf{Z}^{\star}-\text{diag}(\mathbf{Z}^{\star}))_{j} \right\|_{q}\\
\leq &\lambda_{3}\sum_{j\notin \mathcal{D}}\left\|(\mathbf{V}^{\star}-\text{diag}(\mathbf{V}^{\star}))_{j} \right\|_{q}+
\lambda_{5}\sum_{j\in \mathcal{D}}\left\|(\mathbf{V}^{\star}-\text{diag}(\mathbf{V}^{\star}))_{j} \right\|_{q}
+\frac{\lambda_3}{2}\sum_{j=1}^p \left\|(\mathbf{Z}^{\star}-\text{diag}(\mathbf{Z}^{\star}))_{j} \right\|_{q}\\
\leq &\lambda_{3}\sum_{j\notin \mathcal{D}}\left\|(\mathbf{V}^{\star}-\text{diag}(\mathbf{V}^{\star}))_{j} \right\|_{q}+
\lambda_{5}\sum_{j\in \mathcal{D}}\left\|(\mathbf{V}^{\star}-\text{diag}(\mathbf{V}^{\star}))_{j} \right\|_{q}
+\frac{\lambda_3}{2}\left\|\mathbf{Z}^{\star}-\text{diag}(\mathbf{Z}^{\star}) \right\|_{1},\\
\end{split}
\end{equation*}
where the last inequality follows from the fact that for any vector $\mathbf{x}\in \mathbb{R}^p$ and $q\geq 1, \left\|\mathbf{x}\right\|_q$ is a non-increasing function of $q$.

Sum up the two inequalities above and we have
\begin{equation*}
\begin{split}
&\lambda_{1} \left\|\hat{\mathbf{Z}}-\text{diag}(\hat{\mathbf{Z}}) \right\|_{1}+\lambda_{2}\sum_{j\notin \mathcal{D}}\left\|(\hat{\mathbf{V}}-\text{diag}(\hat{\mathbf{V}}))_{j} \right\|_1
+ \lambda_{4}\sum_{j\in \mathcal{D}}\left\|(\hat{\mathbf{V}}-\text{diag}(\hat{\mathbf{V}}))_{j} \right\|_1\\+ &\lambda_{3}\sum_{j\notin \mathcal{D}}\left\|(\hat{\mathbf{V}}-\text{diag}(\hat{\mathbf{V}}))_{j} \right\|_{q}
+\lambda_{5}\sum_{j\in \mathcal{D}}\left\|(\hat{\mathbf{V}}-\text{diag}(\hat{\mathbf{V}}))_{j} \right\|_{q}\\
\leq &\lambda_2 \sum_{j\notin \mathcal{D}}\left\|(\mathbf{V}^{\star}-\text{diag}(\mathbf{V}^{\star}))_{j} \right\|_1+\lambda_{4}\sum_{j\in \mathcal{D}}\left\|(\mathbf{V}^{\star}-\text{diag}(\mathbf{V}^{\star}))_{j} \right\|_1\\
+&\lambda_{3}\sum_{j\notin \mathcal{D}}\left\|(\mathbf{V}^{\star}-\text{diag}(\mathbf{V}^{\star}))_{j} \right\|_{q}+\lambda_{5}\sum_{j\in \mathcal{D}}\left\|(\mathbf{V}^{\star}-\text{diag}(\mathbf{V}^{\star}))_{j} \right\|_{q}
+\frac{\lambda_2+\lambda_3}{2}\left\|\mathbf{Z}^{\star}-\text{diag}(\mathbf{Z}^{\star})\right\|_1\\
< &\lambda_2 \sum_{j\notin \mathcal{D}}\left\|(\mathbf{V}^{\star}-\text{diag}(\mathbf{V}^{\star}))_{j} \right\|_1+\lambda_{4}\sum_{j\in \mathcal{D}}\left\|(\mathbf{V}^{\star}-\text{diag}(\mathbf{V}^{\star})_{j} \right\|_1\\
+&\lambda_3\sum_{j\notin \mathcal{D}}\left\|(\mathbf{V}^{\star}-\text{diag}(\mathbf{V}^{\star}))_{j} \right\|_{q}+\lambda_{5}\sum_{j\in \mathcal{D}}\left\|(\mathbf{V}^{\star}-\text{diag}(\mathbf{V}^{\star}))_{j} \right\|_{q}
+\lambda_1\left\|\mathbf{Z}^{\star}-\text{diag}(\mathbf{Z}^{\star})\right\|_1,\\
\end{split}
\end{equation*}
which gives a contradiction.

From the two theorems above, here comes a corollary.

{\bf Corollary:} Under constraints $\lambda_4 \leq \lambda_2$ and $\lambda_5 \leq \lambda_3$, if $\mathbf{Z}^{\star}$ is non-diagonal and $\exists j\notin \mathcal{D},i \ne j$, such that $\mathbf{V}^{\star}_{ij} \ne 0$, then $$\frac{\lambda_2}{2}+\frac{\lambda_3}{2(p-1)^{\frac{1}{s}}}\leq\lambda_1\leq\frac{\lambda_2+\lambda_3}{2}.$$

\subsection{ADMM Algorithm for DHGL}

Recall the convex optimization problem (\ref{Eq:general}).

 Very similar to the work by Tan et al. (2014), let $\mathbf{B}=(\mathbf{\Theta},\mathbf{V},\mathbf{Z})$, $\tilde{\mathbf{B}}=(\tilde{\mathbf{\Theta}},\tilde{\mathbf{V}},\tilde{\mathbf{Z}})$,
 \begin{equation*}
\begin{split}
f(\mathbf{B}) &= \ell(\mathbf{X},\mathbf{\Theta})  + \lambda_1 \| \mathbf{Z} - \text{diag}(\mathbf{Z})\|_1 +\lambda_2 \sum_{j\notin\mathcal{D}} \| (\mathbf{V}  - \text{diag}(\mathbf{V}))_j \|_1 + \lambda_3 \sum_{j\notin\mathcal{D}} \| (\mathbf{V}  - \text{diag}(\mathbf{V}))_j \|_q \\
&+ \lambda_4 \sum_{j\in\mathcal{D}} \| (\mathbf{V}  - \text{diag}(\mathbf{V}))_j \|_1 + \lambda_5 \sum_{j\in\mathcal{D}} \| (\mathbf{V}  - \text{diag}(\mathbf{V}))_j \|_q\\
\end{split}
\end{equation*}
and
\[
g(\tilde{\mathbf{B}})=\begin{cases} 0 & \text{if } \tilde{\mathbf{\Theta}} = \tilde{\mathbf{V}}+\tilde{\mathbf{V}}^T + \tilde{\mathbf{Z}}  \\ \infty & \text{otherwise}.\end{cases}
\]

Then, we can rewrite (\ref{Eq:general}) as
\begin{equation}
\label{Eq:reformulate}
\underset{\mathbf{B},\tilde{\mathbf{B}}}{\text{minimize  }} \left\{f(\mathbf{B})+g(\tilde{\mathbf{B}})\right\} \qquad \text{subject to } \mathbf{B}=\tilde{\mathbf{B}}.
\end{equation}

Firstly, we derive the scaled augmented Lagrangian for (\ref{Eq:reformulate})
 \begin{equation*}
\begin{split}
L(\mathbf{B},\tilde{\mathbf{B}},\mathbf{W}) &= \ell(\mathbf{X},\mathbf{\Theta})  + \lambda_1 \| \mathbf{Z} - \text{diag}(\mathbf{Z})\|_1 +\lambda_2 \sum_{j\notin\mathcal{D}} \| (\mathbf{V}  - \text{diag}(\mathbf{V}))_j \|_1  \\
&+ \lambda_3 \sum_{j\notin\mathcal{D}} \| (\mathbf{V}  - \text{diag}(\mathbf{V}))_j \|_q + \lambda_4 \sum_{j\in\mathcal{D}} \| (\mathbf{V}  - \text{diag}(\mathbf{V}))_j \|_1 \\& + \lambda_5 \sum_{j\in\mathcal{D}} \| (\mathbf{V}  - \text{diag}(\mathbf{V}))_j \|_q+g(\tilde{\mathbf{B}})  +\frac{\rho}{2}\|\mathbf{B}-\tilde{\mathbf{B}}+\mathbf{W}  \|^2_F,\\
\end{split}
\end{equation*}
where $\mathbf{B}$ and $\tilde{\mathbf{B}}$ are the primal variables, and ${\bf W}=({\bf W}_1, {\bf W}_2, {\bf W}_3 )$ is the dual variable.

Then same as Tan et al. (2014), we use the updating rules:
\begin{enumerate}
\item $\mathbf{B}^{(t+1)} \leftarrow \underset{\mathbf{B}}{\text{argmin  }} L(\mathbf{B},\tilde{\mathbf{B}}^{t},\mathbf{W}^{t})$,
\item $\tilde{\mathbf{B}}^{(t+1)} \leftarrow \underset{\tilde{\mathbf{B}}}{\text{argmin  }} L(\mathbf{B}^{(t+1)},\tilde{\mathbf{B}},\mathbf{W}^{t})$,
\item $\mathbf{W}^{(t+1)} \leftarrow \mathbf{W}^{t}+\mathbf{B}^{(t+1)}-\tilde{\mathbf{B}}^{(t+1)}$.
\end{enumerate}

The updating details are very similar to those shown in Appendix A of Tan et al. (2014), where updates for $\mathbf{B}$ are essentially the same, and updates for $\tilde{\mathbf{B}}$ are exactly the same. So we have the ADMM algorithm for solving (\ref{Eq:general}) as below. Notice that comparing with Algorithm 1 in Tan et al. (2014), the only difference is updates for $\mathbf{V}$. Since we use tuning parameters $\lambda_2$ and $\lambda_3$ for $j\notin\mathcal{D}$, and $\lambda_4$ and $\lambda_5$ for $j\in\mathcal{D}$, we update columns of $\mathbf{V}$ with index $j\notin\mathcal{D}$ and $j\in\mathcal{D}$, separately, using the same form but different tuning parameters.

\begin{algorithm}[htp]
\small
\caption{ADMM Algorithm for Solving (\ref{Eq:general}).}
\label{Alg:general}
\begin{enumerate}
\item  \textbf{Initialize} the parameters:
\begin{enumerate}
\item primal variables $\mathbf{\Theta,V,Z}, \tilde{\mathbf{\Theta}},\tilde{\mathbf{V}}$, and $\tilde{\mathbf{Z}}$ to the $p \times p$ identity matrix.
\item dual variables $\mathbf{W}_1,\mathbf{W}_2$, and $\mathbf{W}_3$ to the $p \times p$ zero matrix.
\item  constants $\rho>0$ and $\tau>0$.\\

\end{enumerate}

\item  \textbf{Iterate} until the stopping criterion $\frac{\| {\mathbf{\Theta}}_{t}- {\mathbf{\Theta}}_{t-1} \|_F^2}{\| {\mathbf{\Theta}}_{t-1}\|_F^2} \le \tau$ is met, where ${\bf \Theta}_t$ is the value of $\bf \Theta$ obtained at the $t$th iteration:

\begin{enumerate}
\item Update ${\bf \Theta}, {\bf V}, {\bf Z}$:
\begin{enumerate}
\item $\mathbf{\Theta}= \underset{\mathbf{\Theta}\in \mathcal{S}}{\arg \min} \left \{  \ell(\mathbf{X},\mathbf{\Theta}) + \frac{\rho}{2} \|\mathbf{\Theta}-\tilde{\mathbf{\Theta}}+\mathbf{W}_{1}\|_F^2 \right \}$.
\item $\mathbf{Z}= S(\tilde{\bf Z} - \mathbf{W}_3, \frac{\lambda_1}{\rho})$,
diag$\mathbf{(Z)}= \text{diag}(\tilde{\mathbf{Z}}-\mathbf{W}_3)$. Here $S$ denotes the soft-thresholding operator, applied element-wise to a matrix:  $S(A_{ij},b) = \text{sign}(A_{ij}) \max( |A_{ij}|-b, 0)$.

\item $\mathbf{C}= \tilde{\mathbf{V}}-\mathbf{W}_2-\text{diag}(\tilde{\mathbf{V}}-\mathbf{W}_2)$.
\item  $\mathbf{V}_j= \max \left(1-\frac{\lambda_3}{\rho \|S(\mathbf{C}_j,\lambda_2/\rho) \|_2}, 0 \right) \cdot S(\mathbf{C}_j, \lambda_2/\rho)$ for $j\notin\mathcal{D}$.
\item  $\mathbf{V}_j= \max \left(1-\frac{\lambda_5}{\rho \|S(\mathbf{C}_j,\lambda_4/\rho) \|_2}, 0 \right) \cdot S(\mathbf{C}_j, \lambda_4/\rho)$ for $j\in\mathcal{D}$.
\item diag$(\mathbf{V}) = \text{diag}(\tilde{\mathbf{V}}-\mathbf{W}_2)$.
\end{enumerate}

\item Update   $\tilde{\bf \Theta}, \tilde{\bf V}, \tilde{\bf Z}$:
\begin{enumerate}
\item $\mathbf{\Gamma} = \frac{\rho}{6}\left[ ({\mathbf{\Theta+W}_1}) - (\mathbf{V+W}_2) -(\mathbf{V+W}_2)^T - (\mathbf{Z+W}_3)   \right]$.

\item $\tilde{\mathbf{\Theta}}= {\bf \Theta + W}_1 - \frac{1}{\rho}\mathbf{\Gamma}$;   \;  \;\;  iii. $\tilde{\mathbf{V}} = \frac{1}{\rho} (\mathbf{\Gamma+\Gamma}^T)  +\mathbf{V+W}_2$;   \;  \;\;  iv. $\tilde{\mathbf{Z}} = \frac{1}{\rho} \mathbf{\Gamma} + {\bf Z+W}_3$.

\end{enumerate}

\item Update  $\mathbf{W}_1, {\bf W}_2, {\bf W}_3$:
\begin{enumerate}
\item $\mathbf{W}_1 = \mathbf{W}_1 + \mathbf{\Theta}-\tilde{\bf \Theta}$;   \;  \;\;  ii. $\mathbf{W}_2 = \mathbf{W}_2 + \mathbf{V}-\tilde{\bf V}$;   \;  \;\;  iii. $\mathbf{W}_3 = \mathbf{W}_3 + \mathbf{Z}-\tilde{\bf Z}$.
\end{enumerate}

\end{enumerate}
\end{enumerate}
\end{algorithm}

Finally, we use Algorithm 1 to solve DHGL. Here we have $\ell(\mathbf{X},\mathbf{\Theta}) = -\log \det  \mathbf{\Theta} + \text{trace}(\mathbf{S\Theta})$. The only thing needed to be specified additionally is the update for $\mathbf{\Theta}$ in Step 2(a)i, which is exactly the same as the update in Tan et al. (2014). We firstly perform eigen-decomposition of $\tilde{\mathbf{\Theta}}-\mathbf{W}_1 -\frac{1}{\rho}\mathbf{S}$ by $\mathbf{UDU}^T$, then update $\mathbf{\Theta}$ by
\[
{\mathbf{\Theta}}=\frac{1}{2}\mathbf{U}\left( \mathbf{D}+ \sqrt{\mathbf{D}^2+\frac{4}{\rho}\mathbf{I}} \right) \mathbf{U}^T.
\]

We can calculate the complexity of Algorithm 1 to solve DHGL, which is $\mathcal{O}(p^3)$ per iteration. Notice that it is dominated by the eigen-decomposition mentioned above.


\begin{thebibliography}{0}
\bibitem{Drton}
Drton M., \& Maathuis MH. (2017) Structure Learning in Graphical Modeling. {\it Annual Review of Statistics and Its Application,} Vol. 4.

\bibitem{Tan}
Tan KM., London P., Mohan K., Lee S., Fazel M., \& Witten D. (2014) Learning Graphical Models With Hubs. {\it Journal of Machine Learning Research,} 15: 3297-3331.

\bibitem{Yuan}
Yuan M.,  \& Lin Y. (2007) Model selection and estimation in regression with grouped variables. {\it Journal of the Royal Statistical Society, Series B,} 68:49-67.

\bibitem{Friedman}
Friedman J., Hastie T., \& Tibshirani R. (2007) Sparse inverse covariance estimation with the graphical lasso. {\it Biostatistics,} 9:432-441.

\bibitem{Boyd}
Boyd S., Parikh N., Chu E., Peleato B., \& Eckstein J. (2010) Distributed optimization and statistical learning via the ADMM. {\it Foundations and Trends in Machine Learning,} 3(1):1-122.
\end{thebibliography}
\end{document}